# HSANET: A HYBRID SELF-CROSS ATTENTION NETWORK FOR REMOTE SENSING CHANGE DETECTION

*Chengxi Han[1], Xiaoyu Su, Zhiqiang Wei, Meiqi Hu[2], Yichu Xu[3]\**

[1]Intelligent Science & Technology Academy Limited of CASIC, Beijing 100043, China
[2]School of Geography and Planning, Sun Yat-sen University, Guangzhou 510006, China
[3]School of Computer Science, Wuhan University, Wuhan 430072, China

**ABSTRACT**

The remote sensing image change detection task is an essential method for large-scale monitoring. We propose HSANet, a network that uses hierarchical convolution to extract multi-scale features. It incorporates hybrid self-attention and cross-attention mechanisms to learn and fuse global and cross-scale information. This enables HSANet to capture global context at different scales and integrate cross-scale features, refining edge details and improving detection performance. We will also open-source our model code: https://github.com/ChengxiHAN/HSANet.

## 1. INTRODUCTION

Change Detection (CD) is the process of identifying differences in the state of an object or phenomenon by observing it at different times [1]. CD plays a crucial role in remote sensing interpretation, with applications spanning land use and land cover analysis [2], disaster assessment [4], environmental monitoring [5], and urban expansion studies [6].

Deep learning techniques have significantly advanced change detection in remote sensing. Models such as FC-EF [7], FC-Siam-conc [7], and FC-Siam-diff [7] utilize stacked convolutional layers to extract robust, discriminative features, surpassing traditional methods. However, these models often exhibit issues such as detection holes and errors. To address these challenges, researchers have incorporated dilated convolutions and attention mechanisms to expand the receptive field and capture global information. Notable models employing these techniques include HCGMNet [8], STANet [9], CGNet [10], SNUNet [11], C2FNet [12], MSPSNet [13], and HANet [14]. Despite these advancements, these models still face challenges in capturing subtle details and addressing neglected areas.

In the realm of remote sensing (RS), including change detection (CD), transformers have become increasingly popular. Models such as BIT [15], Change Former [16], and RSP-BIT [17] have been introduced to achieve a more effective receptive field and deliver competitive performance. The recently introduced ChangeMamba [18], a Mamba-based method, is also noteworthy. Nonetheless, the majority of transformer-based models demand substantial computational resources and exhibit slower training speeds, which can hinder their practical application and impede further exploration by researchers.

To address these challenges, we introduce HSANet, a network that employs hierarchical convolution to extract multi-scale features. By integrating hybrid self-attention and cross-attention mechanisms, HSANet learns and fuses global and cross-scale information. This approach enables HSANet to capture global context at various scales and effectively integrate cross-scale features.

## 2. METHODOLOGY

### 2.1. HSANet

As shown in Fig. 1, we propose a novel network architecture, HSANet, specifically designed for change detection in high-resolution remote sensing imagery. The network introduces significant innovations. The architecture is built around multi-scale feature extraction, spatio-temporal information interaction, and feature optimization, which collectively improve the accuracy and robustness of change detection.

Firstly, HSANet uses the hierarchical convolution structure to extract the features of the input multi-temporal remote sensing images (T1 and T2). By extracting multi-scale features layer by layer, the model can effectively capture spatial information from local to global, which lays a solid foundation for subsequent spatio-temporal fusion. Different from traditional single-scale methods, HSANet has higher expressive power in multi-scale modeling.

Secondly, the network employs a hybrid attention mechanism, including self-attention and cross-attention. The self-attention module focuses on extracting global context information within a single temporal image and capturing the correlations between different regions in the image. The cross-attention module focuses on temporal change features and learns the spatio-temporal interaction information between different temporal images. These two attention mechanisms are integrated by the Fusion Module, which enables the efficient fusion of spatio-temporal information and significantly improves the model's ability to understand complex change patterns.

In addition, HSANet further optimizes feature representation by introducing the HSC-AFM (Hierarchical

Scale-aware Feature Module). The HSC-AFM module can sense the differences between multi-scale features, strengthening key features while suppressing redundant information. This mechanism not only effectively enhances the feature representation of the change region but also preserves target edge details, greatly improving the accuracy of change detection.

Through these innovative designs, HSANet can comprehensively capture multi-scale and spatio-temporal change features, demonstrating excellent change detection ability in complex scenes. Especially in weak change and small target detection, the network shows greater robustness, providing an efficient and accurate solution for high-resolution remote sensing image analysis.

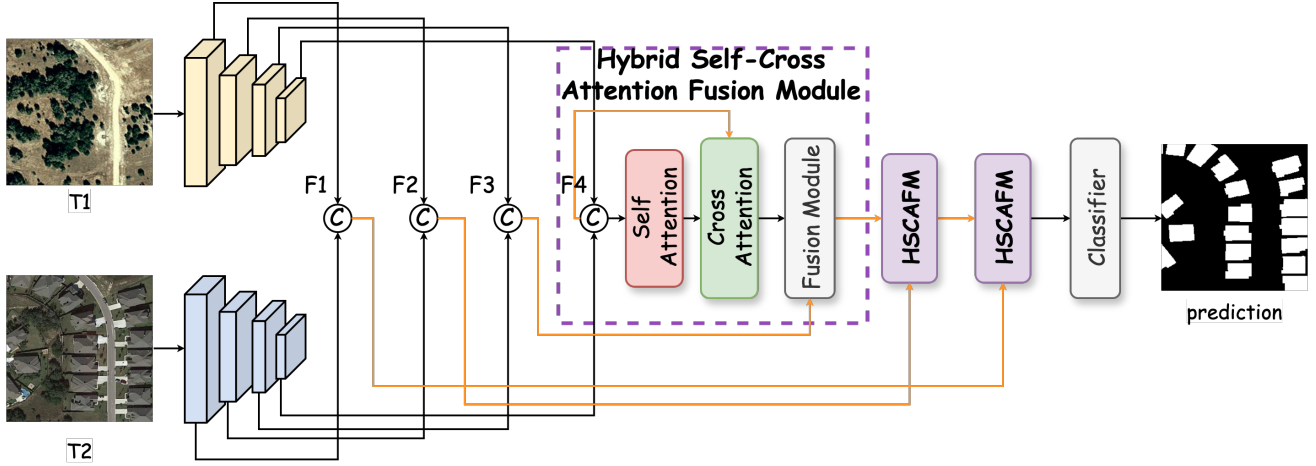

**Fig. 1.** Diagram of the proposed HSANet model.

### 2.2. Hybrid Self-Cross Attention Fusion Module (HSC-AFM)

As shown in Fig.2, HSC-AFM stands for Hybrid Self-Cross Attention Fusion Module, which is the core module in HSANet. First, the input data is passed through a convolutional layer which uses filters (kernels) to extract features. The convolution operation can be expressed as follows.

$$F' = F * W + b \quad (1)$$

Where $F$ is the input feature map, $W$ is the convolution kernel, $b$ is the bias term, $*$ represents the convolution operation, and $F'$ is the output feature map. After the convolutional layers, a Softmax layer is used to assign weights to the features. The features are then passed through two linear layers that integrate the features and inform decisions for the change detection task. The operations of the linear layer can be expressed as follows.

$$Y = W_{linear}X + b_{linear} \quad (2)$$

Where $X$ is the input feature, $W_{linear}$ is the weight matrix, $b_{linear}$ is the bias term, and $Y$ is the output feature. Between the two linear layers, there is a skip connection, which allows the network to pass information directly from the convolutional layer to the later layers. This structure facilitates the propagation of gradients and reduces the problem of vanishing gradients. After the skip connection, there is an addition operation, which adds the output of the convolutional layer to the output of the linear layer, which helps the network to learn a richer feature representation. On the right side of the figure, there is a separate convolutional block, which may represent a more complex convolution operation for a specific feature extraction or processing task. The design of this module allows it to efficiently extract and process features in change detection tasks, capturing spatial features through convolutional layers, integrating features through linear layers, and enhancing the expressiveness of features through skip connections and addition operations. This structure helps to improve the accuracy and robustness of change detection.

### 2.3. Loss Function

Our loss function is the dice loss function, which is calculated as follows.

$$DiceLoss = 1 - \frac{2\sum_{i=1}^{N} y_i \hat{y}_i}{\sum_{i=1}^{N} y_i + \sum_{i=1}^{N} \hat{y}_i} \quad (3)$$

where $y_i$ and $\hat{y}_i$ denote the ground truth label and the predicted value for pixel $i$. $N$ represents the total number of pixels, calculated as the product of the number of pixels per image and the batch size.

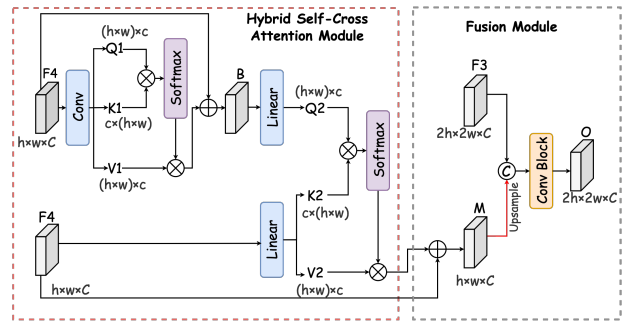

**Fig. 2.** Diagram of the proposed HSC-AFM.

### 3. EXPERIMENTAL SETUP

### 3.1. Datasets

We evaluate our proposed model on two publicly available change detection (CD) datasets: WHU-CD [19] and LEVIR-CD [9]. For the WHU-CD dataset, we divided it into non-overlapping patches of size 256 × 256, resulting in 4536, 504, and 2760 patch pairs for training, validation, and testing, respectively. Similarly, the LEVIR-CD dataset was partitioned into non-overlapping patches of the same size, producing training, validation, and testing sets containing 7120, 1024, and 2048 samples, respectively.

### 3.2. Implementation Details

Our models are implemented using PyTorch and trained on a single NVIDIA RTX 3090 GPU. We utilize the AdamW optimizer with a weight decay of 0.0025 and an initial learning rate of 5e-4 to optimize the loss function. The batch size is set to 8, and the training process spans 50 epochs.

### 3.3. Performance Metrics

To facilitate a more intuitive comparison, we evaluate the performance of our model against SOTA methods using metrics such as F1-score (F1), Precision (Pre.), Recall (Rec.), Overall Accuracy (OA), and Intersection over Union (IoU). These metrics are derived by comparing the ground truth with the predicted maps.

### 4. RESULTS AND DISCUSSION

In this section, we evaluate the change detection (CD) performance of our HSANet by comparing it with several state-of-the-art methods: the stacked convolutional layers methods (FC-EF [7], FC-Siam-conc [7], and FC-Siam-diff [7]), attention mechanisms methods( STANet [9], SNUNet [11], MSPSNet [13] and HANet [14] ) and the transformers-based methods( BIT [15], Change Former [16], and RSP-BIT [17]) .

TABLE I  QUANTITATIVE COMPARISON.

| Model | LEVIR-CD | | | | | WHU-CD | | | | |
|---|---|---|---|---|---|---|---|---|---|---|
| | F1 | Pre. | Rec. | OA | IoU | F1 | Pre. | Rec. | OA | IoU |
| FC-EF [7] | 61.52 | 73.31 | 53.00 | - | 44.43 | 58.05 | 76.49 | 46.77 | - | 40.89 |
| FC-Siam-conc [7] | 64.41 | **95.30** | 48.65 | - | 47.51 | 63.99 | 72.06 | 57.55 | - | 47.05 |
| FC-Siam-diff [7] | 89.00 | 91.76 | 86.40 | - | 80.18 | 86.31 | **89.63** | 83.22 | - | 75.91 |
| STANet-PAM [9] | 85.20 | 80.80 | **90.10** | 98.40 | 74.22 | 82.00 | 75.70 | 89.30 | 98.60 | 69.44 |
| SNUNet [11] | 89.97 | 91.31 | 88.67 | 98.99 | 81.77 | **87.76** | 87.84 | 87.68 | 99.13 | **78.19** |
| MSPSNet [13] | 89.67 | 90.75 | 88.61 | 98.96 | 81.27 | 86.49 | 87.84 | 85.17 | 99.05 | 76.19 |
| HANet [14] | **90.28** | 91.21 | 89.36 | **99.02** | **82.27** | **88.16** | 88.30 | 88.01 | **99.16** | **78.82** |
| BIT [15] | 89.94 | 90.33 | **89.56** | 98.98 | 81.72 | 80.97 | 74.01 | **89.37** | 98.51 | 68.02 |
| Change Former [16] | **90.20** | **92.05** | 88.37 | **99.01** | **82.21** | 87.18 | **92.70** | 82.28 | **99.14** | 77.27 |
| RSP-BIT [17] | 89.71 | 92.00 | 87.53 | 98.98 | 81.34 | 78.50 | 69.93 | **89.45** | 98.26 | 64.60 |
| **HSANet (Ours)** | **91.96** | **93.27** | **90.68** | **99.46** | **85.11** | **92.06** | **94.02** | **90.18** | **99.45** | **85.29** |

* ALL VALUES ARE IN %. FOR CONVENIENCE: **BEST**, **2ND-BEST**, AND **3RD-BEST**.

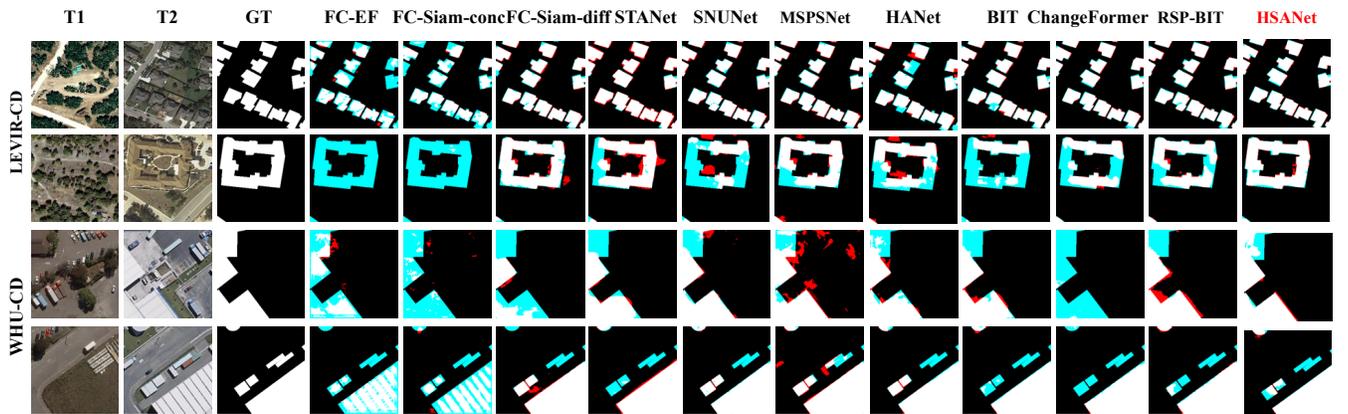

**Fig. 3.** Visual qualitative comparison. For clarity: TP (true positive, shown in white), FP (false positive, shown in red), TN (true negative, shown in black), and FN (false negative, shown in blue).

As shown in Tab. I, the HSANet model exhibits excellent performance. On the LEVIR-CD dataset, the F1 score of HSANet reaches 91.96, precision is 93.27, recall is 90.68, Overall Precision (OA) is as high as 99.46, and Intersection Over Union (IoU) is 85.11, all of which rank first among the compared models. These results demonstrate that HSANet has significant advantages in accuracy, recall rate, and overall classification performance in change detection. In particular, HSANet excels in terms of precision and IoU, meaning it can identify change regions with high accuracy, and the recognition overlaps with the actual change regions are very high. On the WHU-CD dataset, HSANet also performs well, with an F1 score of 92.06, precision of 94.02, recall of 90.18, OA of 99.45, and IoU of 85.29. These results further confirm

HSANet's leading position in the change detection task. The high accuracy and IoU values are especially noteworthy, as they not only reflect the model's precision in identifying changed regions but also demonstrate the high alignment between its predictions and the actual changed regions. Considering the performance on both datasets, the advantages of HSANet in the field of change detection are evident. Its superior performance across key metrics not only proves the effectiveness and reliability of HSANet for change detection tasks but also supports its potential for practical applications. These strengths make HSANet a strong candidate for both change detection research and real-world applications.

As shown in Fig. 3, the HSANet model performs particularly well in change detection. On the LEVIR-CD dataset, HSANet accurately identifies the changed regions (white regions) while maintaining low false positive (red regions) and false negative (blue regions) examples. On the WHU-CD dataset, HSANet also shows high accuracy, with detection results closely matching the true changes (GT), and it has the fewest false positive and false negative cases among all models. This indicates that HSANet has a significant advantage in reducing false positives and false negatives. Compared with other models, such as FC-EF, FC-Siam-conc, FC-Siam-diff, STANet, SNUNet, MSPSNet, HANet, BIT, ChangeFormer, RSP-BIT, and others, HSANet demonstrates clear advantages in both accuracy and robustness in change detection. This further confirms HSANet's high performance in quantitative comparisons; on both the LEVIR-CD and WHU-CD datasets, HSANet achieves the highest scores in key indicators, such as F1 score, precision, recall, overall accuracy, and intersection-over-union ratio. These visual results are consistent with the previous quantitative analysis and further highlight the superiority of HSANet in change detection.

## 5. CONCLUSION

In this paper, we introduce HSANet, a network that employs hierarchical convolution to extract multi-scale features. By integrating hybrid self-attention and cross-attention mechanisms, it learns and merges global and cross-scale information. This approach enables HSANet to capture global context at multiple scales and effectively integrate cross-scale features, thereby refining edge details and enhancing detection performance.